\documentclass[manuscript]{acmart}
\usepackage{multirow}
\usepackage{array}

\usepackage{xcolor}

\newsavebox{\acmboxbox}
\definecolor{mybg}{RGB}{246,247,248} 

\newenvironment{acmbox}[1][\linewidth]{%
  \par\smallskip\noindent
  \setlength{\fboxrule}{0.2pt}%
  \setlength{\fboxsep}{4pt}%
  \begin{lrbox}{\acmboxbox}%
    \begin{minipage}{#1}%
      \setlength{\parindent}{0pt}%
}{%
    \end{minipage}%
  \end{lrbox}%
  \noindent\fcolorbox{black!10}{mybg}{\usebox{\acmboxbox}}%
  \par\smallskip\noindent
}

\newenvironment{eqbox}{%
  \begin{acmbox}%
    \setlength{\abovedisplayskip}{3pt}%
    \setlength{\belowdisplayskip}{3pt}%
    \setlength{\abovedisplayshortskip}{3pt}%
    \setlength{\belowdisplayshortskip}{3pt}%
}{%
  \end{acmbox}%
}

\usepackage{framed}

\newenvironment{boxedquote}{%
  \par\smallskip
  \def\FrameCommand##1{%
    \setlength{\fboxrule}{0.2pt}%
    \setlength{\fboxsep}{4pt}%
    \fcolorbox{black!10}{mybg}{##1}%
  }%
  \MakeFramed{%
    \hsize=\dimexpr\hsize-2\fboxsep-2\fboxrule\relax%
    \leftskip=1em%
    \rightskip=1em%
    \FrameRestore%
  }%
  \noindent\itshape%
}{%
  \endMakeFramed
  \par\smallskip
}

\AtBeginDocument{%
  }

\begin{document}

\title{Emergent, not Immanent: A Baradian Reading of Explainable AI}

\author{Fabio Morreale}
\email{fabio.morreale@sony.com}
\author{Joan Serrà}
\email{joan.serra@sony.com}
\author{Yuki Mitsufuji}
\email{yuhki.mitsufuji@sony.com}
\affiliation{%
  \institution{Sony AI}
  % \country{}
}


\begin{abstract}
Explainable AI (XAI) is frequently positioned as a technical problem of revealing the inner workings of an AI model. This position is affected by unexamined onto-epistemological assumptions: meaning is treated as immanent to the model, the explainer is positioned outside the system, and a causal structure is presumed recoverable through computational techniques. In this paper, we draw on Barad’s agential realism to develop an alternative onto-epistemology of XAI. We propose that interpretations are material-discursive performances that emerge from situated entanglements of the AI model with humans, context, and the interpretative apparatus. To develop this position, we read a comprehensive set of XAI methods through agential realism and reveal the assumptions and limitations that underpin several of these methods. We then articulate the framework’s ethical dimension and propose design directions for XAI interfaces that support emergent interpretation, using a speculative text-to-music interface as a case study.
\end{abstract}




\maketitle

\section{Introduction}
Explainable\footnote{Related terms ---in particular, \textit{interpretability} and \textit{transparency}--- have been proposed, and different stakeholders attribute them different meanings \cite{ehsan_social_2024}. Here, we adopt them as functionally equivalent \cite{linardatos_explainable_2020}, as their finer differentiation lies beyond the scope of this work.} AI (XAI) has been the subject of a large amount of multidisciplinary research, including HCI, for decades. Despite this work, the capacity to interpret AI systems continues to be identified as one of the most pressing and unresolved objectives by scholars \cite{ehsan_expanding_2021,sokol_interpretable_2024} and governments alike \cite{the_white_house_winning_2025}. While work focusing on identifying its sociotechnical aspects is growing \cite{ehsan_expanding_2021,ehsan_charting_2023,muller_interpretability_2025,sokol_what_2024}, prevailing discourses tend to frame interpretability as a technical problem of revealing a causal structure \cite{stamboliev_proposing_2023}, which can only be solved through better computational tools or mathematical functions. These discourses align with representationalist, mechanistic, and positivist conceptions of explanation, which are based on three main assumptions: i) the \textit{explanandum} (what needs explaining) exists within the model; ii) (some) \textit{explananda} are discoverable and accessible; and iii) the explanation tools and the human explainer have negligible epistemic effect on the \textit{explanandum}.

While these assumptions are rarely made explicit, they underpin much of mainstream XAI and have sedimented into a vernacular image of AI models as ``black boxes''. This metaphor, as Langdon Winner warned, naturalises obscurity while foreclosing inquiry into the socio-technical constitution of the system \cite{winner_upon_1993}. Bruno Latour similarly argued that technologies become black-boxed through deliberate acts of closure, thus opacity is not a neutral property of systems but the effect of sociotechnical work \cite{latour_science_1987}. Ananny and Crawford further criticised this metaphor for sidelining the material and ideological complexities of \textit{seeing} and for assuming that accountability can be simply achieved by exposing internal workings \cite{ananny_seeing_2018}. Clinging to this metaphor reaffirms an \textit{ontology} in which meaning resides inside the model, prior to and independent of any constitutive practices; an \textit{epistemology} premised on uncovering such hidden meaning; and an \textit{ethics} that equates accountability with transparency. What is needed is an alternative ethico-onto-epistemology that rethinks from the ground-up what AI interpretations are (ontology), how they are generated and understood (epistemology), and they relate to accountability and responsibility (ethics). Several scholars have already raised critical concerns about dominant assumptions in mainstream XAI \cite{ehsan_social_2024,krishnan_against_2020,carboni_eye_2023,alpsancar_explanation_2024,nicenboim_explanations_2022}. However, these contributions are not intended as onto-epistemological readings of the commitments that underpin prevailing notions of interpretation in XAI.

In this paper, we develop such an onto-epistemological reading by drawing on Karen Barad's agential realism framework and the associated epistemic optic of \textit{diffraction}~\cite{barad_posthumanist_2003}, itself inspired by Donna Haraway's work \cite{haraway_promises_1992}. Barad first developed diffraction as a method to diagnose the problematic onto-epistemological assumptions of scientific practice and eventually developed it into a complete alternative metaphysics: agential realism \cite{barad_posthumanist_2003}. Within agential realism, knowledge is not produced by a detached subject observing an independent object, but emerges through situated \textit{intra-actions} in which the subject and the object are mutually constituted. In this view, knowing is a practice of world-making, thus epistemology and ontology are inseparable.  

Agential realism and diffraction have been applied to multidisciplinary research and, increasingly, in HCI \cite{sanches_diffraction--action_2022,reed_shifting_2024,morrison_entangling_2024,mudd_playing_2023,mice_super_2022,robson_thinking_2024,zhu_centers_2025,bird_diffraction_2025,bomba_choreographer-performer_2024,scurto_prototyping_2021,giaccardi_diffractive_2025,nicenboim_conversation_2023} but, to the best of our knowledge, not directly to XAI. An agential-realist reading of XAI challenges the assumptions i--iii listed above, showing that explanations might, after all, not reside within the model, waiting to be discovered and extracted, but rather emerge from situated intra-actions between the AI model, human interpreters, datasets, XAI tools, and the socio-technical context through which they are performed. This reading shifts interpretability from an exercise of uncovering pre-existing explanations to a material-discursive practice in which the entanglements of more-than-human actors participate in the co-constitution of the explanation. To develop this position, we examine a set of representative XAI methods through Baradian optics. Our analysis reveals conceptual shortcomings inherent in the XAI methods that treat explanations as pre-existing entities awaiting discovery by a human observer, while also showing that some of these methods in fact operate in ways that are better described as diffractive.

The main contribution of this paper is a novel onto-epistemolog\-ical reading of XAI in which interpretability is reframed as a material–discursive performance. This contribution intervenes in debates on critical XAI \cite{burrell_how_2016,krishnan_against_2020,carboni_eye_2023,gilpin_explaining_2018,fazi_beyond_2021,alpsancar_explanation_2024,nicenboim_explanations_2022} by understanding interpretability through entanglement-based~\cite{frauenberger_entanglement_2019} and more-than-human~\cite{eriksson_more-than-human_2024,fuchsberger_doing_2025} perspectives common in HCI, which acknowledge that human and non-human actors participate in generating of knowledge and shaping reality, and recognise technologies as co-constitutive rather than neutral. While HCI has long engaged with XAI matters through a heterogeneous body of work~\cite{kim_help_2023,bryan-kinns_xaixarts_2025,panigutti_understanding_2022,weitz_explaining_2024}, particularly in relation to interactive and human-in-the-loop approaches to explanation~\cite{bertrand_selective_2023,bhattacharya_explanatory_2025} and to responsible and socially situated perspectives on XAI~\cite{liao_designerly_2023,liao_ux_2024,ehsan_who_2024,ehsan_new_2025,de_brito_duarte_amplifying_2025}, entanglement theories have not yet been directly brought to bear on interpretability. We made this connection explicit, thereby both advancing XAI and assessing the reach of HCI entanglement frameworks in a new setting. Also, the ethical implication of our proposed framework joins HCI discourse on ethical AI and social justice, showing how explanation practices are entangled with institutional and economic configurations, which diffractive approaches can help surface. In addition to this main contribution, this paper also offers possible design directions for XAI interfaces centred on emergent interpretations, proposing that multiple and even contrasting explanations should be staged and negotiated through situated, ambiguous interfaces. 

This paper is structured as follows. Section 2 introduces the philosophical theories and concepts that ground our work. Section 3 overviews a large number of XAI methods, which are analysed using Baradian optics as epistemic tools in Section 4. The discussions in Section 5 introduce the ethical consequences of our revised understanding of XAI, and Section 6 develops design directions for XAI interfaces, illustrated through a speculative text-to-music case study. Finally, Section 7  draws conclusions, outlines study limitations and discusses future work.

\section{Background}
In this section, we introduce philosophical concepts and scholarly work that ground our work.

\subsection{Positivism and Representationalism}
Science and engineering research, including a substantial portion of current HCI work, inherits from a philosophical tradition of \textit{positivism}, which proposes that an objective reality exists and scientific tools can be used to observe and make sense of it in a value-free manner \cite{smith_decolonizing_2012}.
Relatedly, \textit{representationalism} is a description used by authors opposing the idea that reality is always already mediated through epistemic processes like signs and language, which have ``the power to mirror preexisting phenomena'' \cite{barad_posthumanist_2003}.
This formulation is grounded in the metaphysical separation of the subject and the object \cite{introna_substantialist_2016} and presupposes that there exists a ``distinction between representations and that which they purport to represent'', and that individuals exist ``awaiting/inviting representation'' \cite{barad_posthumanist_2003}. Thus, \textit{beings} exist anterior to, and independent of, their representations, and instrumentation is separate from what they measure \cite{hollin_disentangling_2017}.  
Beginning with second-wave feminism, and carried forward by contemporary feminist scholarship, radical critiques of Western philosophy have challenged its tenets \cite{smith_decolonizing_2012}, including representationalism and positivism, especially in the work of authors like Michael Foucault, Judith Butler, and, in particular, Joseph Rouse and Karen Barad. 

\subsection{Optics of Knowledge Production}
Østerlund and colleagues \cite{osterlund_building_2020} systematised Barad’s use of optical metaphors \cite{barad_posthumanist_2003} into three distinct optics-refractive, reflective, and diffractive—offering a framework for analysing different epistemological stances in knowledge production.

\textbf{Refraction} was only indirectly mentioned by Barad, who grouped it with reflection~\cite{osterlund_building_2020}. However, Østerlund and colleagues suggest that the refractive optic warrants separate treatment. In this optic, knowledge is produced by revealing, through precise observations, pre-existing structures that ``exist out there''. The metaphor at play is that of a \textit{lens}, which offers direct access to stable and observable objects. A refractive approach thus treats the measurement apparatus as neutral. This stance underpins many engineering and computational traditions, and continues to dominate large portions of scientific practice.

\textbf{Reflection} repositions knowledge production as an interpretative act of a reality reflected by a \textit{mirror}. This optics maintains that a reality with an intrinsic structure and pre-given objects exists and is available to an observer, but admits the impossibility of gaining direct access to reality. The observer can only capture distorted or incomplete reflections, thus the produced knowledge ``is not a perfect replica but a partial reconstruction of reality'' \cite{osterlund_building_2020}. In this optic, the apparatus is no longer entirely neutral, but is still separate from the object and the observer. The world remains pre-structured; the observer is now just more aware of their limitations in representing it. The aim is thus still representational, but this optic acknowledges the interpretative work involved in making sense of the observed object. Barad criticises this framework for still \textit{holding the world at a distance} and for not solving the epistemological gap between subject and object, while acknowledging the attempt to “put the investigative subject back in the picture”~\cite{barad_meeting_2007}. 

\textbf{Diffraction}, in physics, refers to the bending and spreading of waves when they encounter an obstacle or pass through a narrow opening. A classic demonstration of diffraction is Thomas Young’s double-slit interference experiment from 1803. In this experiment, light passing through two closely spaced slits produces, on a screen, an interference pattern\footnote{The pattern arises from the superposition of the diffracted waves emerging from the slits: where crests meet, the waves reinforce each other (bright), and where a crest meets a trough, they cancel (dark).} of alternating bright and dark bands. This phenomenon is also perceivable when a razor blade is illuminated by monochromatic light: its shadow is not sharply delineated, but a pattern of alternating light and dark lines is exhibited~\cite{barad_meeting_2007}. In the twentieth century, similar experiments were repeated with individual photons and other quantum particles. In these versions, the appearance or disappearance of the interference pattern depends on whether the experimenter attempts to measure which slit the particle passes through. It was in this context that physicist Niels Bohr introduced his principle of complementarity ~\cite{bohr_quantum_1928}, arguing that quantum entities cannot be fully captured by either a purely particle-like or a purely wave-like description: what is observed depends on the measuring apparatus. Bohr’s work on complementarity provided the scientific grounds for challenging representationalism. Barad extended this insight beyond physics, mobilising it into metaphysics to articulate the concept of diffraction, which she later developed into agential realism.

\subsection{Agential Realism}
Agential realism is Barad’s broader onto-epistemological framework, of which diffraction is one methodological optic. Agential realism posits that i) there is no absolute separation between observer and observed; ii) measurements do not \textit{represent} ``measurement-independent states of being''~\cite{barad_posthumanist_2003}; and iii) observations are not transparent windows onto reality. Rather, the observer and the observation/measurement apparatus are part of an entanglement that co-determines, and thus cannot be separated from, what is observed. These entities thus do not exist prior to their relations, but are configured by and materialise through their entanglement; they \textit{intra-act} \cite{lettow_turning_2017}. 

Rather than assuming interactions between already constituted entities, intra-action emphasises how phenomena take form through relational processes \cite{hill_more-than-reflective_2018}.  In this view, what classical epistemology treats as an ontological separation between an observing subject and an observed object is understood as \textit{phenomena}, which are ontological primitive relations that do not presuppose pre-existing \textit{relata} \cite{barad_meeting_2007}. Intra-actions address the Cartesian subject–object distinction by producing an \textit{agential cut}, which temporarily resolves indeterminacy and marks provisional boundaries between subject and object. Outside any specific intra-action, subject and object remain indeterminate \cite{barad_meeting_2007}. Such cuts are enacted through \textit{apparatuses}, which are not static instruments that measure pre-existing and measurement-independent entities but “specific material practices through which local semantic and ontological determinacy are intra-actively enacted”~\cite{barad_meeting_2007}. Apparatuses co-produce phenomena by enacting these cuts, and thus drawing boundaries in sociomaterial reality \cite{osterlund_building_2020,orlikowski_sociomateriality_2010}, and are exclusionary as they make some aspects visible while rendering others invisible.

Under agential realism, knowledge is \textit{performative}: practices of knowing participate in the creation of reality \cite{carboni_eye_2023} and thus nothing exists before it is produced as a \textit{discursive phenomenon}. Ontology and epistemology thus implicate each other: ``there are no entities and no knowable characteristics outside of practices of knowing them'' \cite{carboni_eye_2023}. Entities like the observer, the apparatus, and the observed are not given in advance but emerge via their entanglement through material–discursive practices. The diffraction pattern offers a metaphor for this process: as in the case of a razor blade illuminated by monochromatic light, sharp boundaries dissolve into interference fringes. Boundaries are thus not fixed in advance but emerge as effects of relational configurations. As Barad writes, ``a diffraction pattern does not map where differences appear, but rather maps where the effects of differences appear.'' The diffraction pattern is thus not a representation of what is, but a trace of what has happened and the configuration that brought the phenomenon into being \cite{barad_meeting_2007}. 

\subsection{Feminist Epistemology and XAI}
Donna Haraway, a foundational figure in feminist science and technology studies, criticised the universality and alleged objectivity of mainstream scientific knowledge as incapable of seeing its own biases \cite{haraway_situated_1988}. Haraway's proposed alternative is a \textit{situated} knowledge, which is the product of specific socially-situated (e.g., historical, geographical, and cultural) circumstances. Situatedness, thus, turns down purportedly objective ``views from nowhere'' in order to focus on ``acquiring knowledge in particular contexts'' \cite{hancox-li_epistemic_2021}
by means of a dialogue between subjects in a hierarchical relationship of observer and observed, as proposed by black feminist epistemology \cite{hill_collins_black_2000}. 

Feminist epistemology has been previously employed to assess the assumptions embedded in AI research \cite{ciolfi_felice_doing_2025,drage_engineers_2024,browne_feminist_2023,toupin_shaping_2023}, but relatively few works have applied feminist lenses to XAI. Hancox-Li and Kumar suggested that XAI research should allow for multiple interpretations  \cite{hancox-li_epistemic_2021} and should refocus on subjugated and marginalised populations, allowing them to participate in AI explanations, when relevant. Similarly, Nicenboim and colleagues \cite{nicenboim_explanations_2022} drew upon feminist and posthuman theories to criticise the very idea of explainability. They suggest that a more apt concept is that of \textit{shared understanding} between humans and non-humans involved in explanations. They also reject the idea of a single, correct explanation and invite to explore multiple co-produced understandings. 

A recent work on XAI for bias reduction borrows concepts from feminist epistemology in critiquing what the authors term \textit{technical XAI} \cite{huang_ameliorating_2022}. They suggest that XAI models can only detect a limited amount of bias because they are only simplified representations of real AI models, and thus ``can detect only those biases that their simplifying assumptions are attuned to''. The authors also argued against the objectivity of XAI models, as they are affected by undiscussed assumptions and unquestioned social values. Finally, Klumbyte and colleagues \cite{klumbyte_explaining_2023,klumbyte_towards_2023} argued that feminist principles can enrich XAI by challenging universal explanations, foregrounding power and accountability, and centring marginalised perspectives through participatory and interactive approaches. Despite these contributions, diffraction and agential realism have not yet been systematically brought to bear on XAI. Conversely, research in HCI has increasingly drawn on these theories.

\subsection{Agential Realism and Diffraction in HCI}
Agential realism and diffraction have primarily informed broader strands of theoretical HCI and, in particular, entanglement HCI \cite{frauenberger_entanglement_2019}, posthuman and more-than-human perspectives \cite{eriksson_more-than-human_2024}, and sociomaterial approaches \cite{orlikowski_sociomateriality_2010}. Barad's work has also been applied in practical ways to a variety of domains, including design research \cite{bird_diffraction_2025}, choreography~\cite{bomba_choreographer-performer_2024}, lived data \cite{sanches_diffraction--action_2022}, environmental data \cite{giaccardi_diffractive_2025}, wellbeing technologies~\cite{zhu_centers_2025} and, particularly, interactions with digital musical interfaces \cite{mice_super_2022,morrison_entangling_2024,mudd_playing_2023,robson_thinking_2024,reed_shifting_2024,scurto_prototyping_2021}. Across these works, agential realism and diffraction have been used to better understand aspects like authorship \cite{bomba_choreographer-performer_2024}, ambiguity \cite{reed_shifting_2024}, unpredictability \cite{mudd_playing_2023}, creativity \cite{bird_diffraction_2025}, and agency \cite{morrison_entangling_2024} in more-than-human interactions, as well as to surface assumptions about what sort of ``humans'' do specific technologies embed \cite{zhu_centers_2025}.

Furthermore, diffraction has informed design and research methods. Giaccardi and colleagues proposed \textit{diffractive interfaces} in the context of forest simulation data \cite{giaccardi_diffractive_2025}. They showed how agent-based models enact agential cuts, privileging particular attributes and measurements, and thereby reducing the more-than-human complexity of forests. Their proposed diffractive interfaces make these cuts explicit and enable stakeholders to move between different cuts and measurement practices, exploring how different ways of measuring and representing forest data bring different realities into being. Also inspired by Barad is Morrison and McPherson's method of \textit{diffractive dialogue} \cite{morrison_entangling_2024}, in which each author responded to a series of discussion prompts by drawing on their own disciplinary background. 

Despite this growing engagement with Baradian work, to the best of our knowledge, our work is the first to develop a systematic Baradian onto-epistemology of more-than-human configurations in XAI.
\section{XAI Methods}
In this section, we review common XAI methods, adapting and complementing the taxonomy proposed by \cite{schneider_explainable_2024} with additional techniques relevant to our analysis. Our aim is not to provide a systematic survey, but to offer an overview of representative approaches.

\subsection{Feature Attribution}
Feature attribution methods typically aim to explain a \textit{single} prediction generated by a neural network by identifying which (input or intermediate) features were most influential for an output. The assumption is that the model’s decisions are determined by certain features, the influence of which is inherent in the model, and that an observer should be able to discover (thereby revealing), at least in part, the model’s internal decision logic. 

\textbf{Saliency maps} aim to discover the pixels (or regions) that strongly affect an image classification task and, thus, show where the model “looked” in the input. This technique belongs to visualisa\-tion-based explainers, which are considered “objective” explanation tools since they rely on the model’s own signals (e.g., gradient values), rather than human input~\cite{linardatos_explainable_2020}. A typical example in computer vision is Grad-CAM \cite{selvaraju_grad-cam_2017}, which aims to highlight the regions of an image that are most important for a specific prediction.

\textbf{Perturbation-based} approaches systematically modify the input (e.g., masking it, adding noise, shifting values) and observe the effect of these modifications on the output. \textbf{Ablation-based} approaches are closely related, but intervene instead within the model itself, removing specific components (e.g., neurons, layers) in order to assess their relative contribution to the model’s predictions.

\textbf{Surrogate models} are simplified models specifically designed for XAI, which approximate the behaviour of more complex models. LIME \cite{ribeiro_why_2016}, a classical example of a surrogate model, constructs local, interpretable models by approximating the original model's behaviour using a small linear model. It assumes that complex models are locally linear, allowing the resulting weights to explain the factors influencing the output. Another example is SHAP.\footnote{The taxonomy we follow \cite{schneider_explainable_2024} includes SHAP \cite{lundberg_unified_2017} among the surrogate models. SHAP, however, is rather an explanation model for a model's behaviour (or an approximation of it), but it does not replace the original model with a surrogate.} SHAP values are reportedly the \textit{unique} solution that satisfies a set of desirable axioms for feature importance and are thus believed to yield a single objectively “correct” attribution of credit to features, implying that a model’s prediction can be decomposed exactly into feature contributions.

\textbf{Decomposition-based methods} break down a model's reasoning process and attribute its predictions to specific input features or intermediate computations. Layer-wise relevance propagation (LRP) is one such method \cite{montavon_layer-wise_2019}, designed to explain which input features support a model's outputs by breaking it down into relevance scores and propagating them backwards through the network. For example, when classifying an image of a dog, LRP decomposes the model's ``dog'' prediction into pixel-level relevance scores, revealing that high-relevance regions (e.g., the dog's ears and nose) most strongly supported the output. The output is a relevance heatmap, which provides pixel/neuron-level scores but no human-readable concepts.

\subsection{Sample-based}
Sample-based methods aim to explain model behaviour by analysing input-output relationships across \textit{multiple} examples. Rather than attributing importance to individual features of a single instance, these methods characterise predictions in relation to several samples.

\textbf{Adversarial example} \cite{goodfellow_explaining_2015} methods perform minimal input perturbations that flip outputs while appearing unchanged to humans, thereby possibly revealing what features are critical for decision-making. The central idea is that if a minimal change can flip a prediction, the modified parts are likely important for that prediction.

\textbf{Counterfactual explanations} identify minimal and semantically meaningful changes to the input that alter the model's output. Rather than unveiling internal model structure, counterfactuals focus on how the outcome would change under small changes to the input  \cite{wachter_counterfactual_2017}. The objective is not to open the model’s black box, but to show how different inputs would lead to different outcomes. The perturbation must be noticeable to a human: for example, the goal might be to modify an input so that an `entry denied' outcome becomes `entry approved'.

\textbf{Contrastive explanations} address the question of why a model produced a specific output instead of a plausible alternative \cite{miller_contrastive_2021}. Rather than justifying the prediction in isolation, this method explains the decision by identifying the critical, discriminative features that differentiate the input from cases that would have resulted in the alternative. 

\textbf{Prototype-based} methods such as ProtoPNet \cite{chen_this_2019} explain predictions by attributing them to explicitly learned \textit{prototypes} (e.g., prototypical image patches) and highlighting the corresponding regions in the input. Explanations are generated by visualising the input patches that activate the most similar prototypes (e.g., “the network classifies this as a clay-colored sparrow because the wing pattern resembles that of this prototype sparrow”). By making its reasoning process dependent on this direct comparison to prototypical parts, ProtoPNet aims to align its decision process with human conceptualisation.

\textbf{Influence functions} techniques are used to estimate which training examples had the greatest impact on a prediction~\cite{koh_understanding_2020}. Instead of re-running the original learning algorithm, the method approximates how the model’s behaviour would change if the weight of a training point was emphasised.

\subsection{Probing-based}
While methods defined in 3.1 and 3.2 are applicable to inputs and outputs, probing-based methods aim to inspect the internal model representations to understand whether they encode specific knowledge. To do so, auxiliary simple classifiers (probes) are trained on those representations to predict interpretable, pre-determined features.

\textbf{Knowledge probing} investigates the linguistic and factual knowledge implicitly encoded within a model's internal representations. Probes are trained on the model's embeddings to predict specific properties, such as syntactic or semantic labels.  For example, BERT \cite{devlin_bert_2019} has been probed to see if information like part-of-speech tags could be recovered directly from its hidden states \cite{tenney_bert_2019}. A probe's high accuracy on a task is interpreted as evidence that the corresponding knowledge is stored at that particular layer of the model.

\textbf{Concept-based probing} starts from human-defined concepts (e.g., colour, gender, sentiment) and tries to map them to \textit{directions} in the embedding space. A key method is TCAV (Testing with Concept Activation Vectors) \cite{kim_interpretability_2018}, which quantifies the influence of a concept on a model's predictions.  In TCAV, a user defines a concept (e.g., stripes) by providing a set of exemplary inputs that contain it (e.g., images of various striped patterns). TCAV learns a direction in the activation space that represents that concept and measures the model's sensitivity to changes along it. For instance, it can quantify how much the prediction \textit{zebra} depends on the concept \textit{stripes}.

\textbf{Neuron activation-based probing} operates on the premise that the activations of individual (or groups of) neurons correspond to interpretable, often human-defined concepts. A prominent example \cite{bills_language_2023} employs a larger model (GPT-4) to automatically generate interpretations of the patterns that activate specific neurons in a model. By analysing the responses of neurons across diverse inputs, this approach aims to map neuronal function to discrete elements of human understanding.

\subsection{Mechanistic Interpretability}
Mechanistic interpretability methods aim to ``understand neural networks’ decision-making processes'' \cite{sharkey_open_2025} and are based on two main assumptions: i) that complex AI systems have similar ontological status to biological organisms, and thus require similar epistemic tools to be investigated \cite{kastner_explaining_2024}; ii) that complex network behaviours can be explained by identifying a simpler algorithm and a sparse subset of the network that can execute it \cite{meloux_everything_2025}.
Two main approaches in mechanistic interpretability can be identified: reverse-engineering and circuits \cite{sharkey_open_2025}.

\textbf{Reverse engineering} approaches decompose the network into components and then attempt to identify their functions, hoping that structures in the hidden activations correspond to those of the neural computation \cite{sharkey_open_2025}. The most intuitive candidates for decomposition, individual neurons and attention heads, turned out not to be suitable, as they are not a “natural unit for human understanding” \cite{bricken_towards_2023} and are \textit{polysemantic}; i.e., they encode more than one meaning. Proponents of this method suggested that polysemanticity might be caused by \textit{superposition}, a hypothesised phenomenon where a neural network represents more independent features than it has neurons or activations.

Researchers sought to address superposition by finding \textit{sparse} decompositions \cite{bricken_towards_2023,ameisen_circuit_2025}; i.e. by decomposing activations into sparsely active features that might correspond to human-interpretable concepts \cite{ameisen_circuit_2025}.
The subsequent stage involves formulating and testing hypotheses about the functional roles of these features and their interactions \cite{sharkey_open_2025}. To map these interactions, researchers use tools like \textit{attribution graphs} to visualise computational pathways and causal relations to highlight patterns thought to be most relevant for explaining network behaviour. To manage complexity, these graphs are typically simplified through pruning.

The \textbf{circuits} method focuses on identifying interpretable subgraphs within a network that appear to implement specific algorithms, behaviours, or capabilities \cite{olah_zoom_2020}. To address the problem of polysemantic neurons, circuits construct a local replacement model in which neurons are replaced with interpretable building blocks (\textit{features}), which are intended to be monosemantic, producing a surrogate subnetwork that approximates the original model \cite{ameisen_circuit_2025}. These replacement models are then studied to trace how outputs are computed. This approach presumes a correspondence between internal structures (which are not metaphorical but are intended as faithful implementations of cognitive subroutines) and meaningful behaviour, treating neural networks as systems whose logic can be rendered in human-legible terms. 

\subsection{Interactive XAI}
Interactive XAI does not refer to a specific method, but to a way of engaging users interpreting XAI models. Rather than receiving static outputs, humans actively shape the explanatory process. These approaches treat interpretability as a dialogue, enabling users to interrogate models, test hypotheses, and refine explanations through iterative interaction and direct manipulation.

\textbf{Interactive explanations} allow users to probe, steer, and co‑con\-struct the criteria for a satisfactory explanation. Research in HCI underscored that interaction significantly enhances model legibility. This is exemplified by systems like Gamut, a visual analytics tool designed to investigate how interactive interfaces can enhance model interpretation~\cite{hohman_gamut_2019}.
A complementary paradigm is \textbf{explanatory debugging}, which is based on a collaborative loop: the system details the reasoning behind its predictions, and the user, in turn, provides corrective feedback, making explanation a continuous dialogue rather than a fixed artefact \cite{kulesza_principles_2015}. Knowledge about how the model works materialises gradually, through the user and the system in tandem, refining it.

\textbf{Interactive counterfactual explanations} allow users to modify an input instance (e.g., an image region or text segment), and observe how the model’s prediction and accompanying explanation change in response. An example is the What-If Tool \cite{wexler_what-if_2019}, which provides an interactive interface for probing models by altering input features and immediately observing the impact on prediction.  

\textbf{Interactive feature exploration} tools provide direct control over explanatory content through a GUI, as in Prospector \cite{krause_interacting_2016} or ModelTracker \cite{amershi_modeltracker_2015}. Users can filter or highlight features based on importance, reducing cognitive load. Similarly, \textbf{interactive detail-on-demand overlays} can reveal additional context (e.g., labelling an image region) only upon user interaction (e.g., hovering or clicking), avoiding upfront information overload.

Finally, \textbf{interactive debugging tools} like AGDebugger \cite{epperson_interactive_2025} allow developers to reset an AI agent to a prior state, effectively forking its execution path to test alternative plans. This supports direct manipulation of agent behaviour, enabling developers to interactively formulate and test hypotheses about agent reasoning under varying instructions.
\section{A Baradian Analysis of XAI methods}
In this section, we deploy the Baradian optics of refraction, reflection, and diffraction (with the broader agential realist framework) to interrogate the onto-epistemological assumptions embedded in XAI methods. These optics are research probes rather than classificatory tools, as our aim is not to rank methods, but to expose distinct facets of their philosophical commitments. To support our reasoning, we also represented each optic in mathematical formulation and with pictorials, which we clustered in Table~\ref{tab:optics}, to provide complementary registers to engage with our propositions. 

\begin{table*}[t]
\centering
\renewcommand{\arraystretch}{1}
\begin{tabular}{m{0.6\textwidth} m{0.35\textwidth}}

\toprule
\multicolumn{2}{c}{\textbf{(a) Refraction}}\\
\cmidrule(lr){1-2}
\begin{eqbox}
\begin{equation}
M' = \phi_A(M),\qquad \phi_A \approx I \label{eq:refr1}
\end{equation}
\begin{equation}
E = \psi_H(M') \label{eq:refr2}
\end{equation}
\end{eqbox}

\footnotesize
$M'$ is the representation of the model $M$ provided by the apparatus $A$ (Eq.~\ref{eq:refr1}), which is assumed transparent ($\phi_A \approx I$). $E$ is the explanation produced by the human $H$ interpreting $M'$ (Eq.~\ref{eq:refr2}).
&
\includegraphics[width=0.95\linewidth]{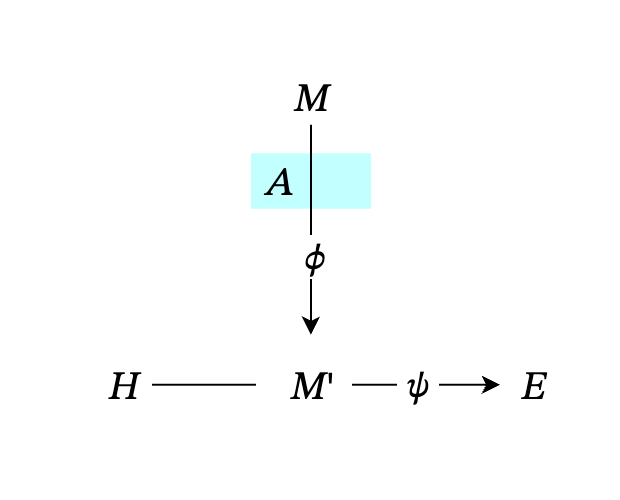}
\\

\midrule
\multicolumn{2}{c}{\textbf{(b) Reflection}}\\
\cmidrule(lr){1-2}
\begin{eqbox}
\begin{equation}
\tilde{M} = \phi_A(M,A),\qquad \phi_A \neq I \label{eq:refl1}
\end{equation}
\begin{equation}
\tilde{E} = \psi_H(\tilde{M}) \label{eq:refl2}
\end{equation}
\end{eqbox}

\footnotesize
$\tilde{M}$ is a reflected/distorted version of $M$ produced by the apparatus $A$ (Eq.~\ref{eq:refl1}). $\tilde{E}$ differs from an ideal explanation $E$ as the apparatus is not transparent ($\phi_A \neq I$) and the interpretation $\psi$ is affected by polarised human optics (Eq.~\ref{eq:refl2}).
&
\includegraphics[width=0.95\linewidth]{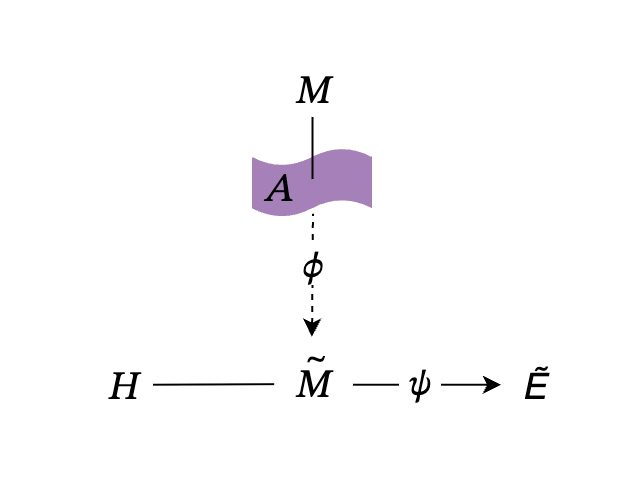}
\\

\midrule
\multicolumn{2}{c}{\textbf{(c) Diffraction / Agential realism}}\\
\cmidrule(lr){1-2}
\begin{eqbox}
\begin{equation}
\delta_i = \mathcal{D}(M, A_i, H_i, C_i) \label{eq:diff1}
\end{equation}
\begin{equation}
\Delta = \sum_{i\in\chi} \delta_i \label{eq:diff2}
\end{equation}
\begin{equation}
E = \Psi_H(\Delta) \label{eq:diff3}
\end{equation}
\begin{equation}
H' = \rho(\Delta, H), \qquad H' \neq H \label{eq:diff4}
\end{equation}
\end{eqbox}

\footnotesize
$\mathcal{D}$ is the entangling operation generating configurations $\delta_i$ from the intra-action of $M$, $A_i$, $H_i$, and different contexts $C_i$  (Eq.~\ref{eq:diff1}). The interference pattern $\Delta$ arises as the structured composition of different $\delta_i$ under the agential cut $\chi$ (Eq.~\ref{eq:diff2}). The human $H$ interprets $\Delta$ to produce the explanation $E$ (Eq.~\ref{eq:diff3}) and dynamically evolves or adapts (Eq.~\ref{eq:diff4}).
&
\includegraphics[width=0.95\linewidth]{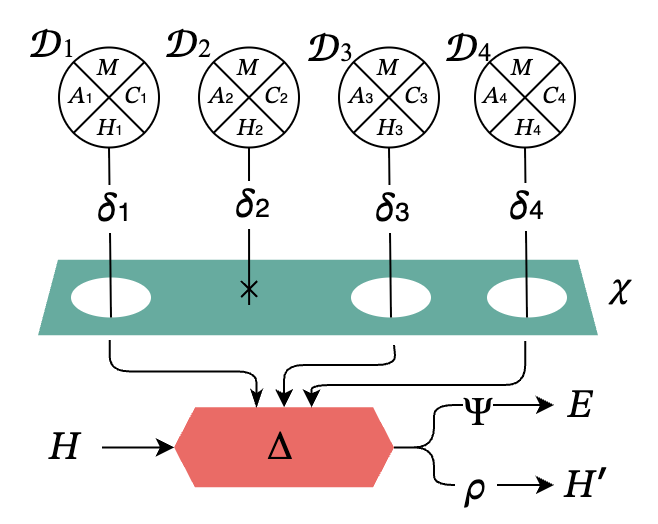}
\\

\bottomrule
\end{tabular}
\caption{Comparison of the three Baradian optics applied to XAI.}
\Description{Table 1 compares three Baradian optics applied to XAI: refraction, reflection, and diffraction / agential realism. Each column gives a compact set of equations and a short verbal and graphical summary that formalise how a model, an apparatus and a human interpreter relate to each other when producing an explanation.}
\label{tab:optics}
\end{table*}

\subsection{Refractive and Reflective Reading of XAI}
In most XAI literature, interpretability is simply treated as an intuitive \textit{desideratum}, but left undefined. In contrast, Erasmus and Brunet characterise interpretation as ``something that one does to an explanation to make it more understandable'' and built a theory of interpretability on top of this definition \cite{erasmus_what_2021}. Here, ``one'' is usually a human expert and the ``doing'' presumes that an explanation already exists, awaiting clarification. 

\subsubsection{Searching Immanent Explananda}
Many XAI methods adopt this stance, treating AI models as if they contain a knowable (discoverable and accessible) essence that they can reveal. This approach evokes an archaeological work, where the task is to uncover structures that are presumed to lie dormant within the model. \textit{Grad-CAM}, for instance, considers that relevance pre-exists in internal feature maps, which are presented as evidence of where the concept (already) resides in the network’s computations. \textit{Circuits}' proponents also aim to recover pre-existing structures and internal mechanisms, and relate to features and connections as real components that ``can be rigorously studied and understood'' \cite{olah_zoom_2020}. Mechanistic interpretability researchers indeed often frame their work as akin to scientists uncovering pre-existing “true statements” about reality, with frequent analogies to biology and neuroscience \cite{ameisen_circuit_2025,lindsey_biology_2025}. \textit{Knowledge probing} tools aim at uncovering some latent linguistic knowledge and thus assume that neurons ``have meanings'' latent within them. As another example, \textit{ProtoPNet}'s architecture is designed to use human-understandable reasoning (prototypes) in its logic, thus assuming that meaning is entirely communicable and interpretable via a set of analogies that humans can relate to.

These methods exemplify refractive and reflective onto-epistemo\-logies as they assume the existence, in AI models, of knowable concepts that exist prior to and outside of the act of interpretation, waiting to be uncovered. We call this an \textit{immanent\footnote{We prefer the term \textit{immanent} over \textit{inherent}, which is more commonly used in XAI literature, since it captures the assumption that explanations are thought to reside within the model itself, as an inner essence waiting to be disclosed, rather than simply being a technical property built into its design.} ontology of XAI}: human-legible \textit{explananda} are already present in the model, and explanation is a technical matter of directly accessing (reflective) or approximating (refractive) them. While they both see the \textit{explananda} as pre-existing interpretations, they differ in how such concepts reach the human observer.  

The \textbf{refractive} reading treats XAI tools as lenses that project sharp images of actual content (Table~\ref{tab:optics}a). \textit{Saliency maps}, for instance, work on the assumption that they can surface “the most meaningful parts” of the image \cite{kim_interpretable_2017}, thus meaning was in the model all along, and the method merely isolates and displays it. \textit{LRP} also sees explanation as an\textit{ exact decomposition} of a model’s output onto input features. The assumption is that the model’s reasoning for a given image exists as a concrete distribution of relevance over the pixels, which LRP neatly uncovers, with no distortion (Eq.~\ref{eq:refr2}). Similarly, \textit{influence functions} assume that predictions are decomposable into quantifiable shares of responsibility assigned to individual training samples. Yet, the predicted explanation is not a causal truth, but a narrative produced through mathematical approximations of model behaviour treated as an objective trace of the model’s computation that the human simply reads.

The \textbf{reflective} reading recognises that XAI methods only approximate pre-existing \textit{explananda} (Table~\ref{tab:optics}b). Thus, in its initial stage, the reflective apparatus operates like a mirror that renders partial or distorted reflections, or approximations, of the AI model (Eqs.~\ref{eq:refl1} and \ref{eq:refl2}). For instance, \textit{SHAP} values provide a simplified account of feature contributions that flattens the actual model's reasoning. Similarly, \textit{perturbation-based} methods approximate feature relevance by selectively masking or altering inputs that supposedly leave the model unchanged. The explanations they produce are \textit{post-hoc} estimates of feature importance, and, as such, they are a map of potential decision pathways, not a transparent window into the model's fixed internal computations.

\subsubsection{Approximating Approximations}
Model approximation is especially pronounced in \textit{ante-hoc} methods as they dismember, simplify, and then recompose the model with human-palatable approximations. Here, the (still pre-existing) object is actively reshaped by the human interpreter prior to its discovery. The interpreter thus not merely observes, but directly \textit{intervenes} in the object of study. Examples include \textit{LIME}, which approximates the \textit{explanandum} with a local surrogate model, and concept erasure models like \textit{LEACE} \cite{belrose_eliciting_2023}, which remove information correlated with certain concepts. In \textit{circuits}, the artificial neurons are replaced with the local replacement model, which is composed of (supposedly interpretable) approximations. These approximations are ``imperfect labels'' \cite{ameisen_circuit_2025} and the model is known to differ from the original model \cite{ameisen_circuit_2025}. Similarly, \textit{attribution graphs} are pruned to only retain nodes and edges deemed to significantly contribute to the output \cite{lindsey_biology_2025}. 

\begin{figure}
    \centering
    \includegraphics[width=0.99\linewidth]{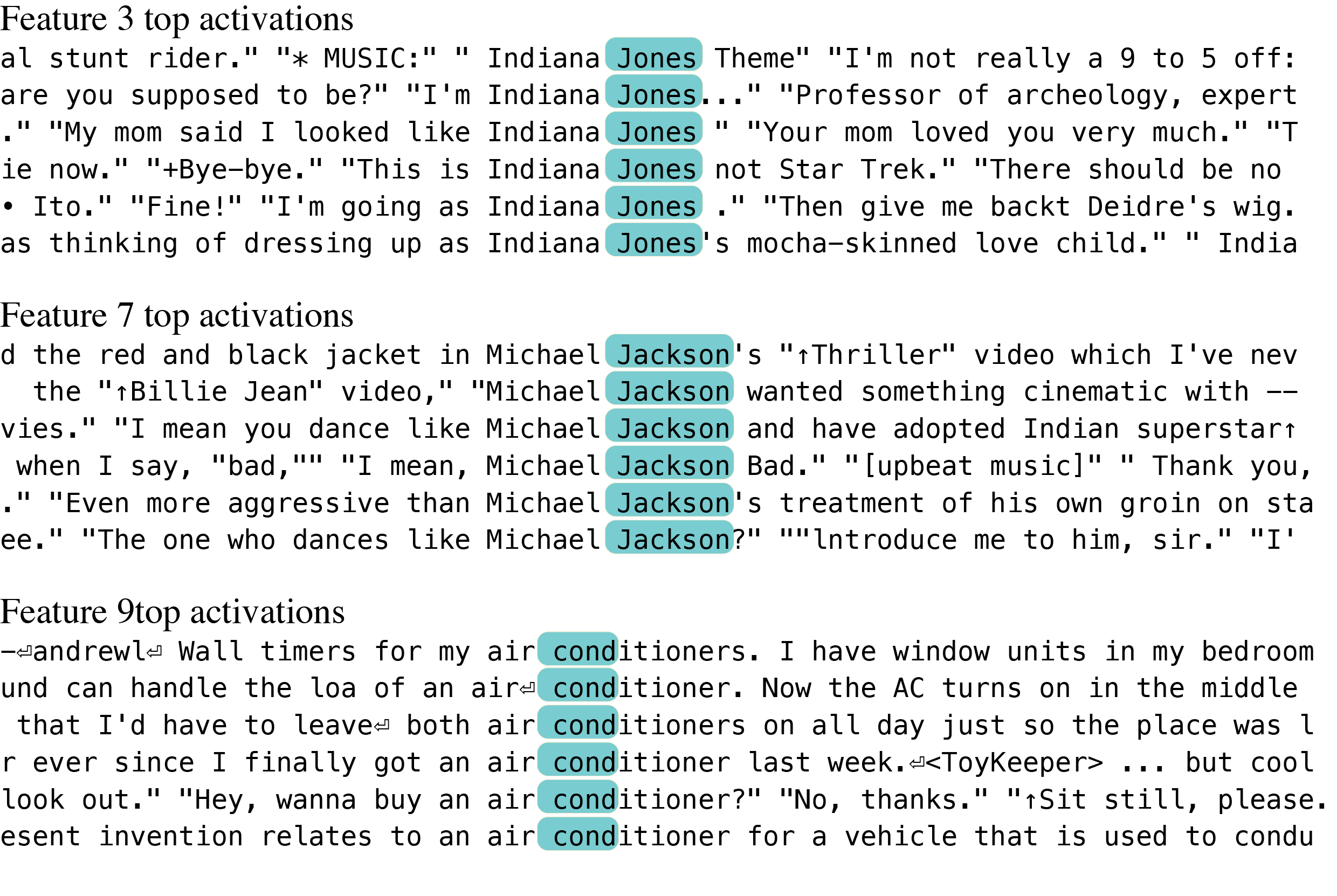}
    \caption{A few of the features grouped into the "Michael Jordan" supernode in \cite{ameisen_circuit_2025}.}
    \label{fig:jordan}
    \Description{Figure 1 shows a grid of feature visualisations that mechanistic interpretability researchers group into a single “Michael Jordan” supernode. The visualisations respond strongly not only to Michael Jordan but also to other people and objects, such as Indiana Jones, Michael Jackson and air conditioners, illustrating that the supernode mixes many heterogeneous patterns rather than a clean, single “Michael Jordan” concept.}
\end{figure}

At the successive stage, an interpretation is required to give meaning to these approximations. The reflected models are here subjected to further distortion when researchers, in the act of interpreting them, impose onto them subjective human optics (Eq.~\ref{eq:refl2}). \textit{Circuits} researchers, for example, group related features into \textit{supernodes} \cite{ameisen_circuit_2025} by identifying human-detectable patterns. These patterns, however, are approximations of approximations and, as such, are prone to errors. For example, one such supernode from~\cite{ameisen_circuit_2025} comprises features that researchers believe activate for “Michael Jordan”; yet, some of these features show stronger activation for Indiana Jones, Michael Jackson, and air conditioners than for Michael Jordan (see Fig.~\ref{fig:jordan}).

The XAI models that operate within these two optics are constructed to identify human-readable and observer-independent properties (e.g., concepts, mechanisms) in a model, even when it has none. Mechanistic explanations, for instance, aim to produce observer-independent \textit{facts} about \textit{how the model works}, assuming that a neural network’s complex behaviour can be reduced to a simpler, human-legible algorithm implemented by a relatively sparse subset of the network \cite{meloux_everything_2025}. However, this belief might be grounded on fallacious premises, which mechanistic interpretability researchers themselves recognise \cite{sharkey_open_2025}, acknowledgeding that ``many of the most strongly-connected features are hard to interpret'' \cite{ameisen_circuit_2025}, that ``more often than not [...] a sparse set of latents that encode some useful concept of interest do not exist [sic]. It is unclear what causes this problem. One hypothesis is that the concept we want isn’t how the model ‘thinks’ about the concept'' \cite{sharkey_open_2025} and that their methods struggle telling apart \textit{faithful} from \textit{plausible} explanations \cite{sharkey_open_2025}.\footnote{This limitation aligns with Carboni and colleagues' comment of \textit{fauxomation}, the false sense of explanation staged by XAI techniques, which ``often offload ambiguity back to human users'' \cite{carboni_eye_2023}.}

\subsubsection{Ante-hoc methods: the Death of the Model}
The onto-espite\-mological implications of \textit{ante-hoc} methods deserve special attention. The simplified/reduced/pruned models are not merely \textit{surrogates}, as they are often framed, but, in Baudrillan's terms, \textit{simulacra} \cite{baudrillard_simulacra_1994}: copies without an original. They are plausible, human-friendly projections obtained through layered reductionist operations (abstraction, decomposition, pruning, selection). These simulacra pretend to represent the model’s logic, yet their grounding is tenuous. These methods offer the comfort of recognising familiar meaning in an alien, possibly unexplainable system, and thus function as epistemological palliatives, structured ways of soothing the desire for clarity in the face of ungraspable inhuman complexity. 
However, like the ethnologist in Baurdillard's story, whose observations cause their object (indigenous people) to vanish \cite{baudrillard_simulacra_1994}, the AI model also ``distintegrates immediately upon contact'' because in ``order for ethnology to live, its object must die; by dying (it) defies the science that wants to grasp it''. In the case of ante-hoc methods, the act of forcing the model into a simplified, interpretable form erases the original system, leaving behind only a projection supposedly tailored to human understanding. The model vanishes, and what remains interpretable is not the model’s actual logic but the residue of the reductionist operations that made it legible. Some mechanistic interpretability researchers are aware of what they call \textit{mechanistic faithfulness}; i.e., the epistemic gap between the replacement and the original model: ``we cannot guarantee that the replacement model has learned the same mechanisms'' \cite{ameisen_circuit_2025}. These methods are indeed based on ``convenient assumptions'', like ``that only active features are involved in the mechanism underlying a model’s responses'' \cite{ameisen_circuit_2025} or on self-identified “speculative claims and assumptions about neural networks” like that “features can be rigorously studied and understood” \cite{olah_zoom_2020}.

\subsubsection{Human-machine Incommensurability}
The various theoretical issues we identified with the XAI methods discussed in this sections might not be easily solved with a technical fix as they might instead stem from the fundamentally different grounds that machines and humans have, which are not directly accessible to the other, what Fazi \cite{fazi_beyond_2021} calls \textit{incommensurability}. Fazi explains that human and machine abstractions are incommensurable: they ``cannot be measured against each other or compared by a common standard''. She contends that XAI aims to bring ``what is beyond human knowledge back into the domain of human cognitive representation'', but warns that ``opening the black box'' might reveal ``nothing to translate or to render precisely because the possibility of human representation never existed in the first place''. Fazi argues that the way in which AI learns features is indeed entirely and exclusively computational and, thus, we cannot have full epistemological access to it. Alvarado echoes this sentiment, suggesting that humans cannot have epistemic access to features that are epistemically relevant for machines, and thus ``explainability is simply not possible'' \cite{alvarado_ai_2023}. Similarly, Burrell identified the flaws in any attempt to impose ``a process of human interpretive reasoning on a mathematical process of statistical optimization'' \cite{burrell_how_2016}. 

These numerous onto-epistemological issues and tenuous assumptions motivate us to look beyond reflectivity and refractivity and to assess whether a diffractive\footnote{As explained in Section 2.3, diffraction, as introduced by Barad, is not simply another optic alongside refraction and reflection but the guiding practice through which her broader framework of agential realism is articulated. Thus, when we speak of an agential-realist onto-epistemology of XAI, we mobilise both the optic of diffraction and the agential realist commitments it entails: intra-action, apparatuses, agential cuts, and the performativity of knowledge practices.} optic offers a more philosophically robust reading of XAI practices.

% ////////////////////////////////////////
% ////////////////////////////////////////
% ////////////////////////////////////////
% ////////////////////////////////////////

\subsection{A Diffractive Reading of XAI within Agential Realism}
In agential realism, knowledge does not pre-exist the practices and the beings that produce it. Similarly, in an agential realism reading of XAI, interpretation is not a unidirectional transfer of pre-existing meaning, but a relational co-production of interpretable \textit{phenomena}. These phenomena emerge through intra-actions between entanglements of human and non-human agencies and actants. This idea resonates with what Ehsan and colleagues \cite{ehsan_expanding_2021} have previously argued about an explanation not being merely a function of the AI system, but a shared meaning-making process situated within social practices. The ability to explain does not reside solely “in the box,” but emerges from who is around the box and what tools they have. On this view, interpretations are \textit{emergent} rather than immanent. 

\subsubsection{Performing Emergent Explanations}
Under agential realism, XAI is a \textit{material-discursive} performance (Table~\ref{tab:optics}c): it involves technical tools (e.g., methods, approximations, visualisation tools) as well as the narratives and contexts that constitute what is made intelligible and what is excluded from intelligibility. A \textit{heatmap} explanation for an image classifier, for instance, is a material artefact (pixels highlighted on a screen) that only becomes meaningful through discourse: “the model focuses on this region because it has learned to associate these pixels with the class \textit{cat}”. The discursive aspect involves attending to such explanations as emerging through different entangled apparatuses (the tools and methods in their situated configuration, Eq.~\ref{eq:diff1}). 

Each interpretive apparatus enacts an agential cut ($\chi$ in Eq.~\ref{eq:diff2}). For example, in \textit{interactive debugging} and \textit{explanatory debugging}, this cut is realised through explicit interaction loops: users inspect outputs and adjust inputs or parameters, so that interpretations arise from the mutual dependence between user actions, model behaviour, and interface representations. \textit{Counterfactual perturbations} enact cuts in which human interpreters are folded into the explanatory practice as alternative inputs are reasoned with by them. Similarly, \textit{contrastive explanations} are situated: the very question of ``why this rather than that'' positions the human interpreter within the explanation, foregrounding what makes one explanation possible against possible alternatives and thus aligning the model’s distinctions with contrasts that the interpreter can perceive and evaluate. Although often presented as a method of locating linguistic knowledge within a model, \textit{probing} can be seen as an agential cut that selects certain features (syntax, roles, entities) as salient while excluding other possible relationalities, and constructs significance by mapping researcher-defined linguistic categories onto the model’s internal representations.

\subsubsection{Analysing Diffractive Signatures}
A diffractive stance stages these readings together and treats interpretation as \textit{reading patterns of difference} through one another (Eqs.~\ref{eq:diff2} and~\ref{eq:diff3}). Brought into deliberate juxtaposition, explanation traces interfere with each other and produce possible interpretations. This juxtaposition can be read through opposition theory, which holds that meaning arises from contrasts rather than from intrinsic properties \cite{saussure_cours_1916}: what a feature or explanation ``is'' becomes intelligible through the differences it makes across different interpretations. Divergences and alignments co-produce an \textit{interference pattern} of intelligibility, where some interpretations are reinforced while others are cancelled and candidate explanations gain credibility where different methods repeatedly converge. The resulting \textit{diffractive signature} is contained in a patterned interrelation of situated explanations rather than in any single one. What counts as an explanation (Eq.\ref{eq:diff3}) therefore emerges through the configuration of model, method, interface, and observer. For instance, a \textit{saliency map} read through a \textit{perturbation analysis} directs attention to features that neither method would have singled out independently.

\subsubsection{Co-constituting Agencies via Boundary-making Performances}\label{sec:coconstituting}
Agential realism expands the epistemological focus of XAI to include all the agencies constituted in the interpretative performance, not just the model and the explanation. As suggested by Orlikowski and Scott, ``if we take [inseparability] seriously the primary unit for research is not independent objects with inherent boundaries and properties but phenomena materially enacted in practice'' \cite{orlikowski_sociomateriality_2010}. Thus, the \textit{explanandum}, the explainer, and the explaining apparatus do not precede but are constituted through the act of interpretation. The explainer does not look at the system from outside of it but is entangled in it; their role and capacity to explain are also enacted within this practice, through which they become a different kind of knowing subject (Eq.~\ref{eq:diff4}). The apparatus itself is also constituted: there is no stable \textit{edge detector neuron}, but only an observed activation that appears related to edges through prior knowledge and technical conventions.

\subsubsection{Opening to Multiple and Uncertain Interpretations}\label{sec:interpretations}
Diffractive XAI not only legitimises but demands emerging interpretations to be \textit{one among others}, not \textit{the} interpretation as different interpretations can and do emerge from different entanglements. Confirmations for this idea come from Nicenboim and colleagues \cite{nicenboim_explanations_2022}, who explained that individual explanations cannot fit all contexts and thus there is a need to design for multiple interpretations to be produced, and from Meloux and colleagues' empirical finding that multiple and even incompatible explanations can be generated with a single method \cite{meloux_everything_2025}. These entanglements are akin to Bohr’s experimental arrangements: each reveals one facet while obscuring others. Plurality and situatedness of interpretations stand in sharp contrast to traditional positions in XAI, which tend to treat ambiguity and multiplicity as problems to be solved in pursuit of a singular, sanitised account of model behaviour. Mechanistic interpretability research indeed assumes \textit{explanatory unicity}; the hypothesis that a single explanation exists for a given phenomenon \cite{meloux_everything_2025}. 
Interpretability also cannot fully eliminate uncertainty because,  a feature of relational sense-making and not an obstacle to it. As commented by Carboni and colleagues, “opacity and uncertainty are not deficits to be eliminated but effects of relational configurations of humans, data, models, and institutional practices” \cite{carboni_eye_2023}.
\section{Ethical Implications}
The mainstream XAI research paradigm downplays researchers' ethical responsibility, as their role to recover or reveal a model’s pre-existing internal structure seems ethically inconsequential. From a Baradian standpoint, however, the onto-epistemological practices through which XAI itself takes shape have three distinct implications.

First, agential cuts that constitute explanatory performances are \textit{acts of responsibility} \cite{sanches_diffraction--action_2022}, as they participate in making some interpretations possible while foreclosing others~\cite{carboni_eye_2023}. Agential cuts are moments in which specific entanglements of models, data, funders, researchers, and institutions are stabilised so that some interpretations come to matter, while others recede. Responsibility for these cuts must be considered and their consequences made visible \cite{frauenberger_entanglement_2019}. Following Fuchsberger and Frauenberger \cite{fuchsberger_doing_2025}, in more-than-human assemblages, responsibility is not static and distributed, but is performed, relational, fluid, and collectively produced; it is \textit{done}. Doing responsibility in XAI means rejecting the idea that responsibilities can be cleanly distributed to individual components of the explanatory entanglements and revealing the imaginary underlying specific entanglements, the funding flows they follow, and the politics they are aligned with, and the power asymmetries embedded within entanglements. An analysis of entanglements that include mechanistic interpretability, for instance, reveals that the field is currently dominated by a small number of research institutions that set the political agenda, which sees interpretability as a goal to be pursued at all costs, despite researchers from these very same institutions warning that progress in interpretability research has been modest and has not yet demonstrably made AI systems safer \cite{sharkey_open_2025}. Responsibilities indeed begin in the milieus of innovation that make particular technological futures imaginable and viable ~\cite{fuchsberger_doing_2025}. In the case of mechanistic interpretability, for instance, such a milieu can be partially reconstructed by following the funding that flows into this research area, the affiliations of its researchers, who mostly work for companies associated with the TESCREAL bundle~\cite{gebru_tescreal_2024}, and their policy-facing ambitions to ``anticipate dangerous capabilities'' and to ``translate technical progress into levers for governance'' \cite{sharkey_open_2025}. 

Second, XAI researchers are accountable for identifying and correcting the patterns of exclusion caused by their explanatory entanglements. Giraud’s \textit{ethics of exclusion} critiques entanglement theories that fail to identify who is systematically left out of entanglements \cite{giraud_what_2019}. Entanglements do not ensure ethical adequacy so they must be interrogated \cite{hollin_disentangling_2017} to surface which explanation methods and metrics are privileged, which kinds of expertise and reporting venues define what counts as a valid explanation, and which individuals and communities affected by model decisions are marginalised. When these exclusions persist, transparency becomes a hollow demand: visibility does not yield accountability if those affected lack the power to make claims or seek remedies, as critiques of the transparency ideal in algorithmic governance have shown \cite{ananny_seeing_2018}. For communities that already experience epistemic exclusion (i.e., by being prevented from participating in collective meaning-making and epistemic processes \cite{huang_ameliorating_2022}), the risk is that entanglements might exacerbate these issues. Framing exclusion as a matter of accountability therefore re-politicises interpretability by asking who is entitled to receive and contest explanations, who must answer for the consequences of model use, and how these obligations are distributed \cite{stamboliev_proposing_2023}.

Third, agential realism offers new critical tools to challenge the vernacular understanding of transparency. Reducing transparency to a matter of black-box opacity naively simplifies questions of accountability \cite{stamboliev_proposing_2023,ananny_seeing_2018}, as visibility alone cannot reveal the sociotechnical context that conditions model behaviour. From an agential realist perspective, transparency is always partial and situated, and is mediated by the specific entanglements that make \textit{seeing} possible. For instance, documentation and analytic practices such as datasheets for datasets, model cards, and genealogies of data work show that making systems legible is not a simple matter of disclosure but of curating accounts of a system’s life cycle \cite{mitchell_model_2019,scheuerman_datasets_2021,denton_genealogy_2021}. These interventions involve situated decisions about which histories, assumptions, and data lineages to foreground and which to omit, thereby stabilising particular accounts of where data come from and who is implicated in their production \cite{denton_genealogy_2021,morreale_data_2023}. These documentation practices are material-discursive configurations that actively participate in the production of accountability.

A diffractive approach therefore reframes (a) \textit{responsibility} as an ongoing, collective practice of making, tracing, and contesting agential cuts within explanatory entanglements, and (b) \textit{accountability} as the situated reconfiguration of who can see, contest, and demand redress for model behaviour within the apparatus of models, explanations, and institutions.

\section{Diffractive Design for Emergent Interpretations}
While developing a comprehensive set of concrete design prescriptions is beyond the scope of this paper, this section outlines some directions to provide a starting point for designing diffractive interfaces that support emergent interpretations. These directions are informed by several works in HCI, and, in particular, by the aforementioned work on diffractive interfaces \cite{giaccardi_diffractive_2025} and by studies on emergence, which is central in our proposed onto-epistemology. Gaver and colleagues articulated how, in design research, the objectives, procedures, and outcomes often do not follow the execution of a fixed plan but \textit{emerge} through making \cite{gaver_emergence_2022}. Building on this view, more recent HCI work focused on understanding how AI systems are configured and emerge through ongoing relations with artefacts and environments. For example, Nicenboim and colleagues used conversational agents and more-than-human prototypes to stage open-ended situations in which partial, situated, and often mistaken understandings of AI can evolve over time \cite{nicenboim_conversation_2023,nicenboim_unmaking-ai_2024}. In parallel, Ghajargar and Bardzell’s “graspable AI” concept cards act as tangible prompts through which designers’ ideas about AI can iteratively emerge \cite{ghajargar_making_2022}. 

In the remainder of this section, after presenting each design direction, we illustrate it with a practical example: a persona interacting with a speculative interface \cite{dunne_speculative_2013} designed to explain the outputs of a generative text-to-music (TTM) system. The interface allows users to generate songs using the AI model and to query the outputs to develop specific interpretations. In line with our framework, the users can define the configurations of the explanatory entanglements and the generations that they create (e.g., XAI methods, training dataset, explainer's background, model version, prompt). We chose this case study because it draws on our disciplinary expertise and because generative music offers an especially fertile setting for investigating diffractive XAI. Assessing the constitutions of explanatory entanglements in TTM is indeed particularly important because: i) the training data used in these systems are indeed often scraped without consent and encode a narrow, biased canon of music \cite{born_diversifying_2020}; ii) the space of allowed outputs is reduced by the epistemic decisions behind these models ~\cite{morreale_reductive_2025}; and iii) their outputs risk actively displacing human labour ~\cite{morreale_unwitting_2023}.

\subsection{Foreground Multiple Interpretations}
Supported by findings from related work \cite{nicenboim_explanations_2022, meloux_everything_2025}, in Section \ref{sec:interpretations}, we explained how diffractive XAI demands multiple interpretations. Diffractive XAI interfaces should thus refrain from visualising and settling on one correct interpretation and should instead show patterns of difference across several explanatory performances. Diffractive XAI interfaces should not only enable users to select different entanglements and agential cuts, but also stage situated outputs against one another so that they can be experienced in contrast, echoing our earlier point that meaning emerges through comparison rather than from isolated, intrinsic properties.

\begin{figure}
    \centering
    \includegraphics[width=0.99\linewidth]{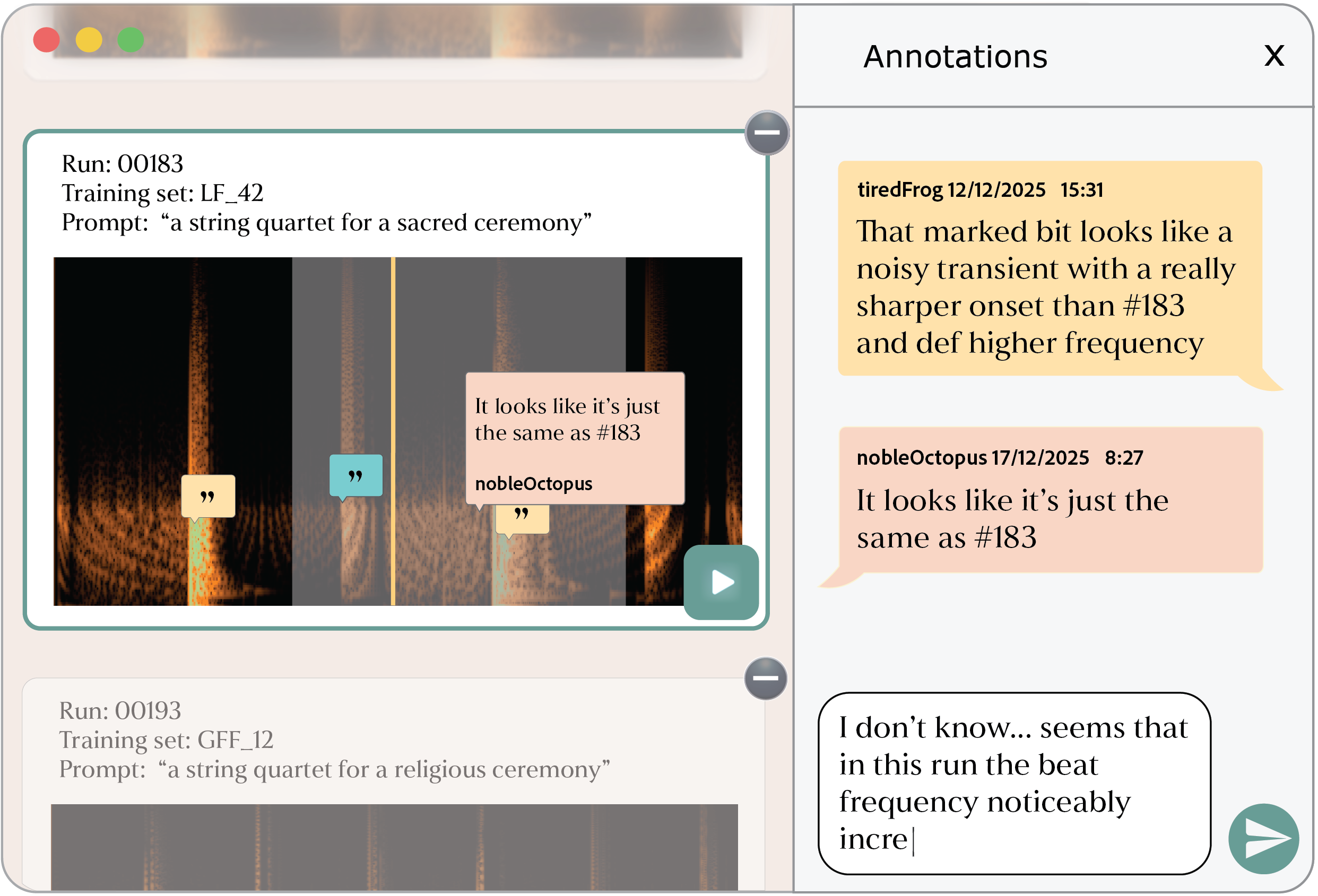}
    \caption{A speculative interface designed to interpret text-to-music generations. It displays multiple situated entanglements annotated by different users.}
    \label{img:xttm}
    \Description{A screenshot of a TTM interface displaying several spectrogram outputs from different model runs, each with its prompt and training-set metadata. The active run panel shows several user comments anchored to regions of the spectrogram. The right sidebar lists three textual annotations discussing transient noise, similarity between runs, and changes in beat frequency. A play button appears below each spectrogram, and a plus icon is visible in the bottom centre for adding new runs.}
\end{figure}

\begin{boxedquote}
A model developer aims to understand how the TTM model has internalised the concept of ``sacred'' in music. They initially fix a specific entanglement and generate a song from a certain prompt (e.g., ``a string quartet for a sacred ceremony''). Then, the model produce counterfactuals by minimally varying components of the entanglement like the prompt (e.g., replacing ``sacred'' with ``religious'') or the subset of the training catalogue (e.g., muting tracks from a given region or label). Different users then annotate how \textit{sacredness} transforms across these versions (Fig. \ref{img:xttm}). An explainer then analyses the interference pattern between these explanatory performances. The pattern might reveal that \textit{sacral} and \textit{religious} co-activate with specific classes of instruments (e.g., organ, choir) that only appear when the training catalogue contains songs from a particular region. The material-discursive performance can thus reveal, for instance, how specific catalogues perpetuate colonial notions of sacral music.
\end{boxedquote}

\subsection{Foreground Situatedness and Repeated Engagement}
In a diffractive view, situatedness and emergence mutually determine interpretations. Explanations arise from partial, situated entanglements and thus cannot be treated in isolation. Rather, they unfold through repeated engagements in which agential cuts shift and new relations become salient. Building on work on diffractive interfaces \cite{giaccardi_diffractive_2025} and practice-based accounts of emergence \cite{gaver_emergence_2022}, diffractive XAI interfaces should 
foreground the situatedness of the entanglements that gave rise to explanations, render agential cuts visible, and allow users to move between them. They should then support revisiting prior explanations and navigating patterns of differences from successive interactions to perform new interpretations. On this view, diffractive XAI interfaces are spaces where explanations continually emerge and re-emerge through changing and evolving relations among the components of the explanatory entanglement.

\begin{boxedquote}
An IP lawyer aims to identify which training song played the greatest role in a specific generation (attribution problem). They perform different generations using the same seed\footnotemark but varying aspects of the entanglements. Initially, they explore different attribution methods (e.g., \textit{influence functions}, and {concept-based probe} for ``compositional style'') and then they select a different subset of the training dataset, and a combination of these. The user can also  analyse candidate songs both visually (by analysing a spectrogram or other visualisation graphs) and acoustically. Each generation-output-explanation cycle is a revisitable interpretive performance related to the specific situatedness of the entanglement. By navigating traces of successive runs as they accumulate over time, and by allowing users to toggle between different entanglements, a candidate explanation emerges.
\end{boxedquote}
\footnotetext{Generative diffusion models often exhibit a ``seed consistency'' property: with a fixed architecture and different but related training data, models can produce very similar outputs when given the same prompt and random seed \cite{kadkhodaie_generalization_2024}.}

\subsection{Foreground Uncertainty and Ambiguity}
Recent work invited to stage misunderstandings and breakdowns in human-AI interaction as occasions for reflection \cite{nicenboim_conversation_2023,nicenboim_unmaking-ai_2024}. This suggestion resonates with long-standing HCI work that treats ambiguity and open meaning as resources for design and interpretation~\cite{gaver_ambiguity_2003,reed_shifting_2024,boehner_advancing_2006,morreale_influence_2019,sengers_staying_2006,wakkary_things_2021} and with the design concept of \textit{seamful design} \cite{chalmers_seamful_2004}, which foregrounds gaps, seams, and incongruities so that multiple readings remain available. Brought to XAI, this principle invites designers to treat uncertainty, ambiguity, non-knowledge, and disagreement as materials for interpretation that should be foregrounded rather than sanitised.

\begin{boxedquote}
A musician aims to identify and make sense of disagreements between various explanations of the outputs of a TTM model. Specifically, they implement \textit{probing heads} to map musical concepts they define (e.g., melody, texture, rhythm) onto the model's internal representations. Rather than averaging these scores to present a single interpretation, the interface deliberately seeks out passages where these probes disagree. For instance, it identifies disagreement where one probe confidently labels a strings sequence as the \textit{main melody}, while another strongly classifies it as \textit{accompaniment}, and a third indicates high activation for \textit{background texture}. The interface foregrounds this conflict and presents these competing interpretations as opportunities for musical inquiry, which might result in the musician redefining the existing concept labels or inventing entirely new ones. In doing so, musicians are not only active participants in the material-discursive practice of interpretation but become a ``different kind of knowing subject'' (Section \ref{sec:coconstituting}), as their musical expertise changes after participating in the explanatory performance.
\end{boxedquote}
\section{Conclusions}

In this paper, we challenged the assumptions of mainstream XAI research about the existence of hidden \textit{explananda} within AI models that can be recovered by an observer who is external to the interpretative process. Inspired by Baradian's agential realism, we proposed an alternative onto-epistemology that reframes interpretability as a performed phenomenon that allows possible, but not definitive, explanations to emerge from an entanglement of models, observers, tools, and contexts. Understanding interpretability as a material-discursive performance co-constituted by situated entanglements of human and non-human actors carries not only onto-epistemic but also ethical stakes. Configuring entanglements is an act of responsibility as each entanglement foregrounds some perspectives while forecloses others. XAI researchers are therefore accountable for these configurations as constitutive conditions that must be made visible, acknowledged, and politicised. XAI research, we argue, might benefit from embracing this onto-epistemological shift: not extracting truths from models, but practicing explanations that remain attentive to difference, interference, and accountability. As for the limitations of this study, our framework is primarily conceptual, and we do not present a full empirical case study of diffractive XAI in use. The design directions we outlined have not been developed into operational design guidelines nor evaluated in fully developed scenarios. Future work should therefore translate these directions into concrete design patterns and guidelines.

\section{Ethical statement}
In this paper, we cited self-published technical works only insofar as they provided insights into their methods and thinking, treating them strictly as secondary data sources. Other non-peer-reviewed literature has not been considered or referenced in order to uphold the standards of academic rigour.

\bibliographystyle{ACM-Reference-Format}
\bibliography{zot_references}

\end{document}